\documentclass[letterpaper,10pt,conference]{ieeeconf}
\IEEEoverridecommandlockouts
\usepackage{times}
\usepackage{cite}
\usepackage{algorithmic}
\usepackage{graphicx}
\usepackage{textcomp}
\usepackage{capt-of}
\usepackage{booktabs}
\usepackage{siunitx}         
\usepackage{threeparttable}  
\let\labelindent\relax
\usepackage{enumitem}
\usepackage{bm} 
\usepackage{makecell}        
\usepackage{amsmath,amssymb,amsfonts}
\usepackage{stfloats} 
\usepackage{soul}     
\usepackage{array}
\usepackage{booktabs}
\usepackage{makecell}

\usepackage{tabularx}
\usepackage{graphicx}
\usepackage[table,xcdraw,dvipsnames]{xcolor}
\usepackage[T1]{fontenc}
\usepackage{epsfig}
\usepackage{amsmath}
\usepackage{amssymb}
\usepackage{color, soul, colortbl}
\usepackage{mathtools}
\usepackage{booktabs}
\usepackage{times}
\usepackage{microtype}
\usepackage[labelfont=bf, font=footnotesize]{caption}
\usepackage[labelfont=bf, font=footnotesize]{subcaption}
\usepackage[utf8]{inputenc}
\usepackage{float}
\usepackage{inconsolata}
\usepackage{multirow}
\usepackage{placeins}
\usepackage{stfloats}
\usepackage{enumitem}
\usepackage{tabularx}
\usepackage{xstring}
\usepackage{multirow}
\usepackage{xspace}
\usepackage{url}
\usepackage{hyperref}
\usepackage{duckuments}
\usepackage{svg}
\usepackage[hang,flushmargin]{footmisc}
\usepackage{url}
\usepackage{tabularray}
\usepackage{wrapfig}
\usepackage{epigraph}
\usepackage{amssymb}

\newcommand{\R}[1]{{%
    \textbf{%
        \ifstrequal{#1}{1}{\textcolor{red}{R#1}}{%
        \ifstrequal{#1}{2}{\textcolor{blue}{R#1}}{%
        \ifstrequal{#1}{3}{\textcolor{magenta}{R#1}}{%
        \ifstrequal{#1}{4}{\textcolor{teal}{R#1}}{%
                           \textcolor{cyan}{R#1}%
        }}}}%
    }%
}}

\definecolor{tabfirst}{rgb}{0.7, 1.0, 0.7} 
\definecolor{tabsecond}{rgb}{1, 1, 0.7} 
\definecolor{tabthird}{rgb}{1, 0.85, 0.7} 

\newcommand{\raisemath}[1]{\mathpalette{\raisemath{#1}}}

\newcommand{\Jc}{\mathcal{J}}

\newcommand{\Wc}{\mathcal{W}}

\newcommand{\Rb}{\mathbb{R}}

\newcommand{\av}{\mathbf{a}}

\newcommand{\gv}{\mathbf{g}}

\newcommand{\ov}{\mathbf{o}}
\newcommand{\pv}{\mathbf{p}}
\newcommand{\qv}{\mathbf{q}}

\newcommand{\sv}{\mathbf{s}}

\newcommand{\xv}{\mathbf{x}}
\newcommand{\yv}{\mathbf{y}}

\newcommand{\Tv}{\mathbf{T}}

\newcommand{\deltav     }{\boldsymbol \delta     }

\definecolor{RebuttalColor}{RGB}{0,0,0}

\definecolor{pmcolor}{RGB}{70,70,70}

\definecolor{checkgreen}{RGB}{34,139,94}
\newcommand{\greencheck}{\textcolor{checkgreen}{\checkmark}}
  
\sethlcolor{yellow}
\usepackage{xspace}
\newcommand{\ours}{\texttt{SimDuoReal}\xspace}

\definecolor{darkblue}{HTML}{234A6F}

\makeatletter
\let\original@makecaption\@makecaption
\long\def\@makecaption#1#2{%
    \ifx\@captype\@IEEEtablestring
        {\footnotesize\normalfont\raggedright\noindent #1: #2\par}%
        \vskip 6pt
    \else
        \original@makecaption{#1}{#2}%
    \fi
}
\makeatother
\usepackage{hyperref}
\usepackage{cleveref}

\definecolor{teaseryellow}{RGB}{225,165,40}

\begin{document}

\title{\LARGE \bf SimDuoReal: Object-Centric Bimanual Policy for Emergent Dexterity
\vspace{2mm}
}

\author{Qiyang Yan$^{1}$, Yulin Liu$^{1,2}$, Akash Sharma$^{1}$,
Haozhi Qi$^{\dagger 1,3}$, Guanya Shi$^{\dagger 1,4}$, C. Karen Liu$^{\dagger 1,5}$\\
$^{1}$Amazon FAR, $^{2}$UCSD, $^{3}$University of Chicago, $^{4}$CMU, $^{5}$Stanford University%
\thanks{$^{\dagger}$Amazon FAR team co-lead.}\\[4pt]
\noalign{\vskip 2mm}
{\hypersetup{hidelinks}\href{https://simduoreal.github.io/}{\textcolor{teaseryellow}{\textbf{simduoreal.github.io}}}}}
\thispagestyle{empty}
\pagestyle{empty}

\let\oldtwocolumn\twocolumn
\renewcommand\twocolumn[1][]{%
\oldtwocolumn[{#1}{%
\centering

\vspace{-3mm}
\includegraphics[width=0.97\textwidth]{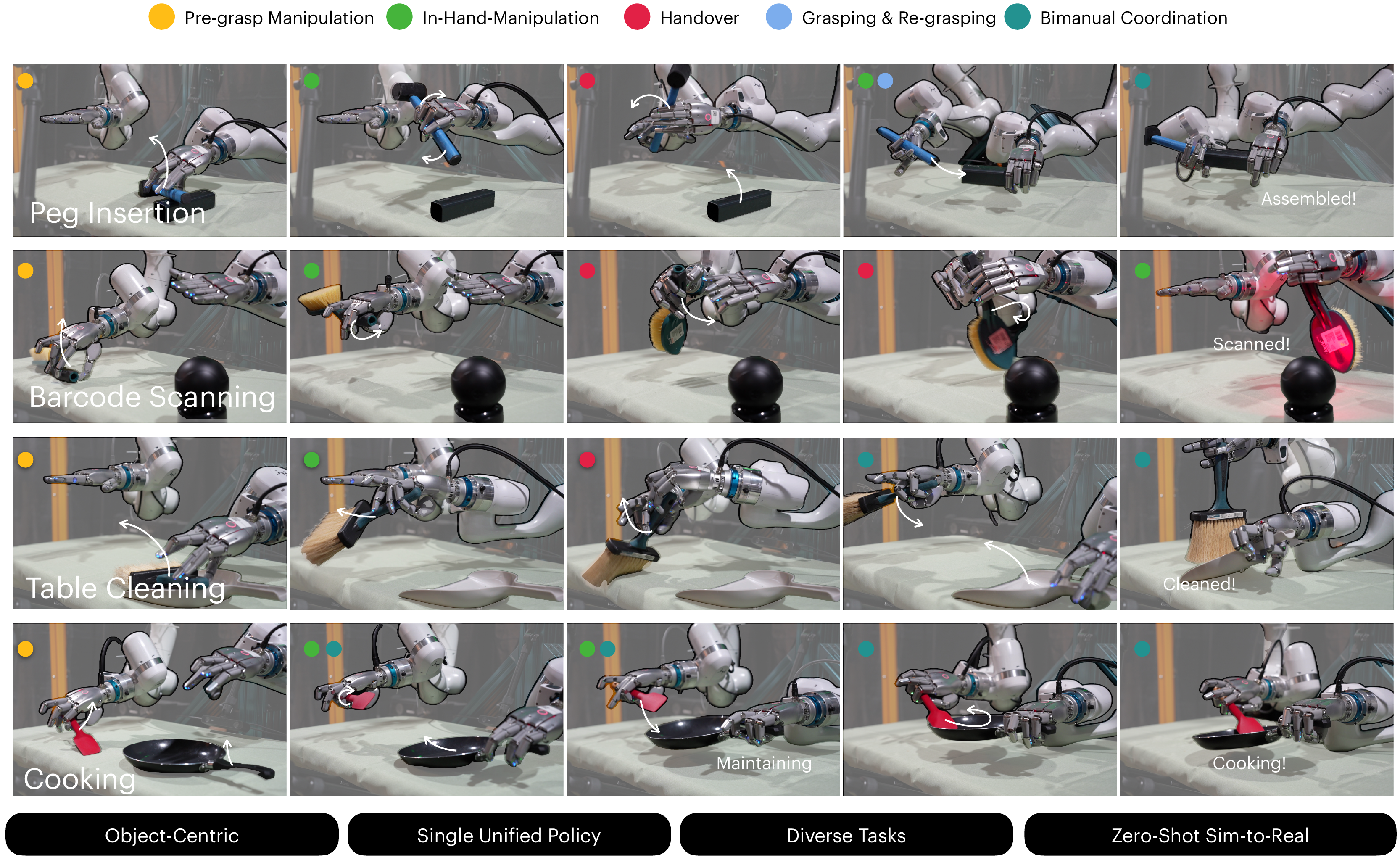}
\captionof{figure}{\small\ours is a single object-centric policy that controls two dexterous arm-hand systems from a stream of desired object configurations. The same policy, transferred zero-shot from simulation, performs assembly, barcode scanning, table cleaning, and cooking on real hardware. Markers denote the behaviors that emerge along each rollout: pre-grasp manipulation (orange), in-hand manipulation (green), handover (red), grasping and re-grasping (blue), and bimanual coordination (teal).}
\label{fig:teaser}
\vspace{3mm}
}]%
}

\maketitle

\begin{abstract}
Bimanual dexterous manipulation requires coordinated arm and hand control, yet specifying such coordination manually is challenging and often task-specific. In this work, we explore whether complex bimanual dexterity can emerge from a simple object-centric learning objective. We train reinforcement learning policies under a shared object-centric reward, varying their training environments rather than tuning separate rewards for each behavior. We find that the policies discover pre-grasp manipulation, grasping and regrasping, in-hand adjustment and handover to reach object-pose goals without hand-designed motion sequences. We then refine the experts on transition states and use DAgger and reinforcement learning to distill them into a single policy that composes these behaviors for diverse bimanual tasks. On real hardware, the same policy succeeds in 61 of 80 trials (76.3\%) across four tasks with unseen objects and no task-specific fine-tuning. In simulated barcode scanning, our policy achieves 85.0\% success, compared with 60.4\% for two single-arm policies without handover. These results suggest that this objective provides a simple and scalable interface for learning general bimanual dexterity, shifting the burden from manually designing robot coordination to specifying what should happen to the object.
\end{abstract}

\section{Introduction}

Bimanual dexterous manipulation systems have the potential to perform a wide range of everyday tasks: picking up an object, reorienting it to expose a desired face, and passing it between hands. Existing systems have demonstrated these capabilities on individual tasks~\cite{kedia2026simtoolreal,andrychowicz2018learning,fraualfaro2025handover}. However, specifying these behaviors is challenging because the robot must decide not only how to move the object, but also which hand to use and when to pass it between hands. Existing methods often specify this coordination through task-specific rewards or skill sequences. These specifications are difficult to reuse across tasks, so extending the system to new behaviors requires repeated engineering.

Object-centric formulations sidestep part of this problem by specifying tasks purely in terms of the object's desired state, leaving the robot free to discover how to bring it about. For example, SimToolReal~\cite{kedia2026simtoolreal} trains a single-arm policy to move procedurally generated tool-like objects to random 6-DoF goal poses and transfers zero-shot to unseen real tools. Tasks can then be specified as sequences of object goal poses without object- or task-specific training. Extending this approach to two hands over a larger workspace, however, poses a learning challenge. For goals beyond either hand's reach, the policy must discover how to transfer the object before it can complete the goal. RL can instead settle for partial progress with one hand without discovering the handover needed to reach the goal.

The straightforward approach is to add task-specific rewards that shape hand selection and handover. This can guide exploration, but requires specifying how the hands should coordinate for each behavior. New behaviors may then require new reward terms, weakening the benefit of a reusable object-goal interface.

In this work, we extend object-centric task specification to bimanual dexterous manipulation by designing \emph{training environments} rather than hand-craft separate \emph{rewards} for each behavior. We vary the distribution of object goals and constrain each hand's workspace under a shared object-centric reward. Bimanual behaviors emerge during training, without explicitly programming how the hands should pick up, reorient, or pass the object (Fig.~\ref{fig:teaser}).

More specifically, we define a reachable workspace for each hand and train two single-arm experts for object control within their respective workspaces. Furthermore, we train a bimanual expert for cross-workspace transfer (Sec.~\ref{sec:method-atomic-skills}). All three share the same object-centric representation and reward, while their training tasks only differ in goal sampling, episode initialization and termination. For single-arm training, the object starts on the table and the goals stay within that arm's reach. The hand learns to pick up the object, reorient it, and change its grasp as the goals change. For handover training, one hand starts holding the object, and the goal is sampled from the other hand's workspace, forcing the holding hand to pass the object to the other hand to reach the goal. A goal is complete only when the receiving hand contacts the specified surface region on the object and the giving hand releases the object. As a result, the giver learns to adjust the object before passing it, exposing this region for a \emph{transfer grasp} rather than a \emph{hold grasp} (Fig.~\ref{fig:qualitative}). We distill the experts into one policy using behavior cloning and RL, after fine-tuning each expert from states reached by preceding experts. 
During distillation, the policy is trained on both individual expert behaviors and their transitions (Sec.~\ref{sec:method-distillation}).

Our policy can be prompted with objects: we specify how the objects should move, and the policy controls the arms and hands to carry out that motion without task-specific training. An operator can provide these prompts by moving virtual objects in VR. We also extract object trajectories from RGB-D human demonstrations using FoundationPose~\cite{wen2024foundationpose} and use them as pose waypoints. We evaluate barcode scanning in simulation with unseen object shapes. On real hardware, we use object-level teleoperation to test unseen objects on four tasks: barcode scanning, peg insertion, cleaning, and cooking (see Fig.~\ref{fig:teaser}). We also compare expert-first and end-to-end training and ablate expert refinement, the distillation loss, and the training scenes (Sec.~\ref{sec:experiments}).

\noindent\textbf{Contributions.}
\begin{itemize}
\item We design training environments that use object-goal sampling and workspace constraints to learn single-arm control and bimanual handover under a shared object-centric reward, rather than separate rewards for each behavior.
\item We show that pickup, in-hand adjustment, regrasping, and handover emerge without hand-designed motion sequences. Before handover, the giver learns \emph{transfer grasps} that expose the object region the receiver needs to grasp.
\item We prompt \emph{one} bimanual policy with object motion to execute new tasks without task-specific training. We demonstrate this with unseen real objects in barcode scanning, peg insertion, cleaning, and cooking.
\end{itemize}

\section{Related Work}

\textbf{Bimanual Dexterous Manipulation.}
Bimanual systems with multifingered hands have demonstrated coordinated
behaviors such as twisting lids, tool use, and passing objects between
hands~\cite{chen2022bidexhands,lin2024twisting,petrenko2023dexpbt,lin2024learning,shaw2024bimanual}.
Reinforcement learning systems specify each behavior through task-specific
rewards or staged sub-goals~\cite{chen2022bidexhands,lin2024twisting,petrenko2023dexpbt}, while
imitation systems specify it through demonstrations collected per
task~\cite{zhao2023aloha,lin2024learning,shaw2024bimanual,jiang2024dexmimicgen}. In
both cases the specification includes which hand acts and when the object
changes hands, so a new behavior requires new reward terms or new
demonstrations. We instead use a single object-centric reward and vary only
the goal distribution and the workspace constraints used during training.

\textbf{Object-Centric Task Specification.}
A second line of work specifies manipulation through object-centric task specification. 
In-hand reorientation reaches 6-DoF object pose goals with a single
hand~\cite{andrychowicz2018learning, handa2023dextreme, chen2022reorientation,yin2023rotating, chen2026ptld},
and recent methods use object trajectories as the learning signal for dexterous
policies~\cite{chen2024objdex,liu2025dextrack,lum2025human2sim2robot}.
SimToolReal~\cite{kedia2026simtoolreal} trains a single-arm policy to move
procedurally generated objects to random goal poses and transfers zero-shot to
unseen real tools, so tasks become sequences of object goal poses. These
formulations assume one hand and one workspace. With two hands, pose error alone
does not distinguish a handover from a single-hand carry, and the carry is
available whenever the goal is reachable; \ours makes the transfer the only
strategy that reaches the goal.

\textbf{Composing Specialists into One Policy.}
Training specialists and compressing them into one network is common practice:
teacher--student distillation folds privileged or terrain-specific policies into
a single deployable one~\cite{rusu2016distillation,ross2011dagger,zhuang2023robot,cheng2024extreme},
and skill-chaining work observes that a policy degrades on the initiation states
its predecessor actually produces, and regularizes terminal states to close that
gap~\cite{lee2021skillchaining,chen2023sequential}. These pipelines still define
each skill by its own reward or reference motion. Our experts differ only in
goal sampling, so one reward supervises both the refinement of each expert on
its predecessors' terminal states and the reinforcement learning term used
during distillation.

\section{Method}

\begin{figure*}[t]
\centering
\includegraphics[width=\textwidth]{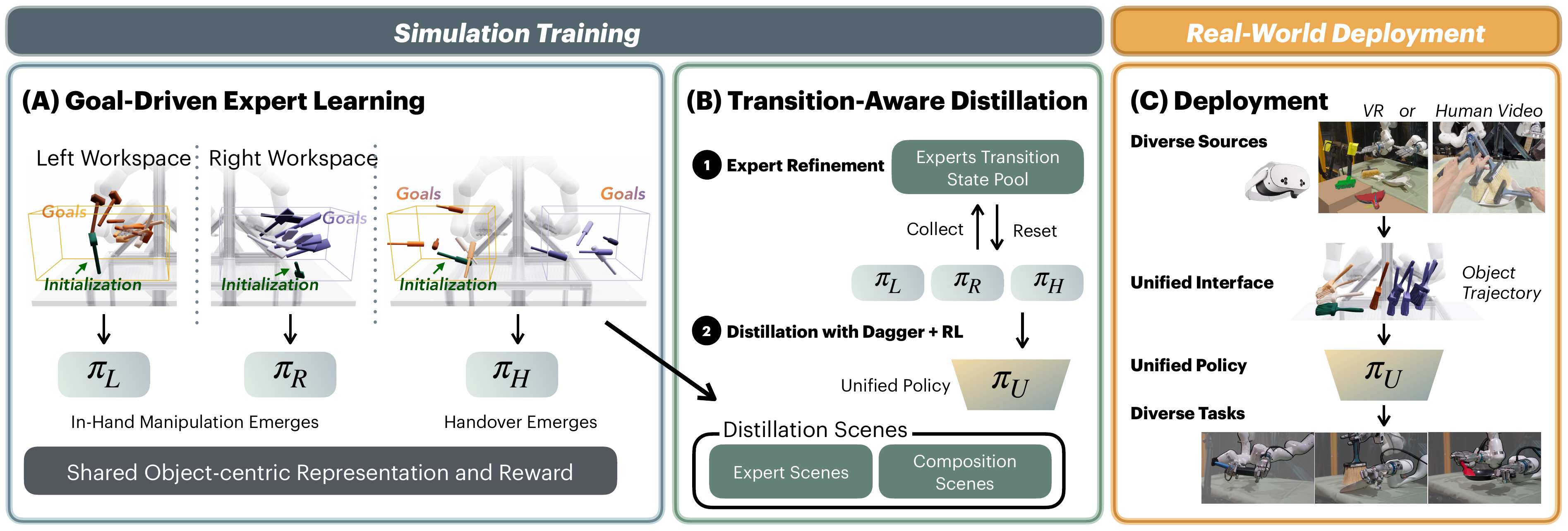}
\caption{\small\textbf{System overview} (A) Three experts are trained in simulation under a shared object-centric representation and reward, with different goal-sampling regimes: the left and right experts $\pi_L$ and $\pi_R$ learn grasping with in-hand manipulation from goals inside one arm's workspace, and the handover expert $\pi_H$ learns handover from goals that shift from the giver's workspace to the receiver's. (B) The experts are refined from a pool of states collected at expert transitions, then distilled into a unified policy $\pi_U$ with Dagger and RL in scenes that contain both individual expert behaviors and their transitions. (C) At deployment, object trajectories from diverse sources such as VR teleoperation and videos serve as the unified interface, and the single policy $\pi_U$ executes diverse real-world tasks.}
\label{fig:overview}
\end{figure*}

We propose \ours, a single bimanual policy for manipulating one or two objects toward desired 6-DoF poses (Fig.~\ref{fig:overview}). Through this common goal interface, \ours acquires dexterous behaviors such as grasping, in-hand adjustment and handover, which can be composed to execute downstream tasks such as coordinated tool use. Our key idea is that these behaviors can emerge from a \textit{shared object-centric} formulation by varying only the training environments (Sec.~\ref{sec:method-atomic-skills}). We first train three workspace-conditioned experts for within-workspace control and cross-workspace transfer. We then refine these experts at transition boundaries and distill them into a unified policy using DAgger and reinforcement learning in scenes spanning individual expert regimes and their sequential or concurrent composition (Secs.~\ref{sec:method-distillation}). At deployment, the unified policy tracks object trajectories obtained from VR or human video (Secs.~\ref{sec:method-prompting}), enabling diverse zero-shot bimanul tool manipulation.

\subsection{Problem Formulation}
\label{sec:method-problem-formulation}

We consider the problem of learning a bimanual dexterous policy that manipulate up to two objects to desired 6-DoF pose sequences. At control step $t$, let $\Jc_t\subseteq\{1,2\}$ index the active objects. For each object $j\in\mathcal{J}_t$, ${\mathbf{T}_t^j,\hat{\mathbf{T}}_t^j}\in \mathbf{SE}(3)$ denote its current and goal poses. The policy observes the robot proprioception $\mathbf{s}_t$, the current and goal object poses, and object-centric geometric representation $\mathbf{g}_t$ introduced in Secs.\ref{sec:method-atomic-skills}: 
\begin{equation}
  \av_t =
  \pi_\theta
  \left(
      \sv_t,
      \{\Tv_t^j,\hat{\Tv}_t^j, \gv_t\}_{j\in\Jc_t}
  \right)
  \in \Rb^{58},
\end{equation}
where $\av_t$ specifies relative displacements from the previous targets for the arms and absolute targets for the hands as ~\cite{kedia2026simtoolreal}.

A task is specified by object-pose goals for each active object, provided online or as waypoint sequences (Sec.~\ref{sec:method-prompting}). The policy is conditioned on the current goals. A goal is complete when the pose error $d_t^j$ falls below $\epsilon$. For cross-workspace object transfer, this condition additionally requires the receiving hand to establish contact and the giving hand to release the object.

\subsection{Learning Workspace Experts through Goal Sampling}
\label{sec:method-atomic-skills}
We define a Left Workspace and a Right Workspace to indicate the reachable space for each hand, with an overlapping zone in the middle (see Fig.~\ref{fig:overview}). We train two single-arm experts $\av_t^{k} = \pi_k(\ov_t^{k}) \in \mathbb{R}^{29}, k \in \{L,R\},$ for object control within their respective workspaces and one bimanual expert $\av_t^{H}=\pi_H(\ov_t)\in\mathbb{R}^{58}$ for cross-workspace transfer. All three share the same object-centric representation and reward, while their training tasks differ in goal sampling and episode initialization and termination. \textbf{\textit{Grasping, in-hand adjustment, and handover}} therefore \textbf{\textit{emerge}} from object-pose reaching without skill-specific motion references or rewards. 

\subsubsection{Goal-Sampling Regimes}
\label{sec:method-goal-sampling}

\paragraph{Within-workspace control
per arm}
The episode begins with an un-grasped object on the table. Following SimToolReal~\cite{kedia2026simtoolreal}, an initial goal position is sampled within the arm's reachable workspace $\Wc_k\subset\Rb^3$, and subsequent goals are generated by translations and rotations while remaining within $\mathcal{W}_k$. Repeatedly reaching these goals requires grasping and sustained object control while also producing \textit{in-hand adjustment and regrasping} as goals change.

\paragraph{Cross-Workspace Transfer}
\label{sec:cross-workspace-goal}
The bimanual expert is trained from states in which one hand, designated as the giver initially holds the object. The goal-sampling region then shifts from the giver's workspace into the receiver's workspace excluding the overlapping zone. We sample $N_{\mathrm{shift}}$ goals in the cross-over setting and $N_{\mathrm{dst}}$ further goals in the destination workspace. We train with both directions by exchanging giver and receiver roles. During expert training we set $N_{\mathrm{shift}}=N_{\mathrm{dst}}=1$ to focus learning on the handover itself. A cross-workspace goal is completed only after the receiver establishes contact and the giver releases the object. Dropping the object or violating the workspace constraints terminates the episode. This goal distribution induces \textit{handover and preparatory object adjustment} without specifying either behavior explicitly.

\subsubsection{Unified Object-Centric Formulation}

\paragraph{Shared Object-Centric Representation}

Building on GraspXL~\cite{zhang2024graspxl}, we extend hand--object displacement features to both hands and add inter-hand displacements for bimanual coordination.
For each object $j$, target and avoidance regions $S_j^+$ and $S_j^-$ specify desired and undesired contact. Let $\xv_{A,i}$ denote the position of keypoint $i$ on the left or right hand, $A\in\{A_L,A_R\}$. Its nearest sampled points in $S_j^+$ and $S_j^-$ are $\yv_{A,i}^+$ and $\yv_{A,i}^-$, respectively. Omitting the object index $j$, the hand--object displacements are
\begin{equation}
\deltav_{A,i}^{+}=\yv_{A,i}^{+}-\xv_{A,i},
\qquad
\deltav_{A,i}^{-}=\yv_{A,i}^{-}-\xv_{A,i}.
\label{eq:object-representation}
\end{equation}
Without a superscript, $\deltav_{A,i}$ denotes the displacement to the nearest sampled point on the full surface $S_j=S_j^+\cup S_j^-$.
Inter-hand displacements are $\deltav_{i,k}^{LR}=\xv_{A_R,k}-\xv_{A_L,i}$ for each pair of left- and right-hand keypoints. Within-workspace experts use only the active hand's hand--object features; the bimanual expert and unified student use both hands' features and inter-hand displacements. We concatenate hand--object features across active objects with inter-hand features where applicable to form $\gv_t$, capturing geometric relationships for coordination regardless of changing contact roles.

\begin{figure}[t]
\centering
\includegraphics[width=0.85\columnwidth]{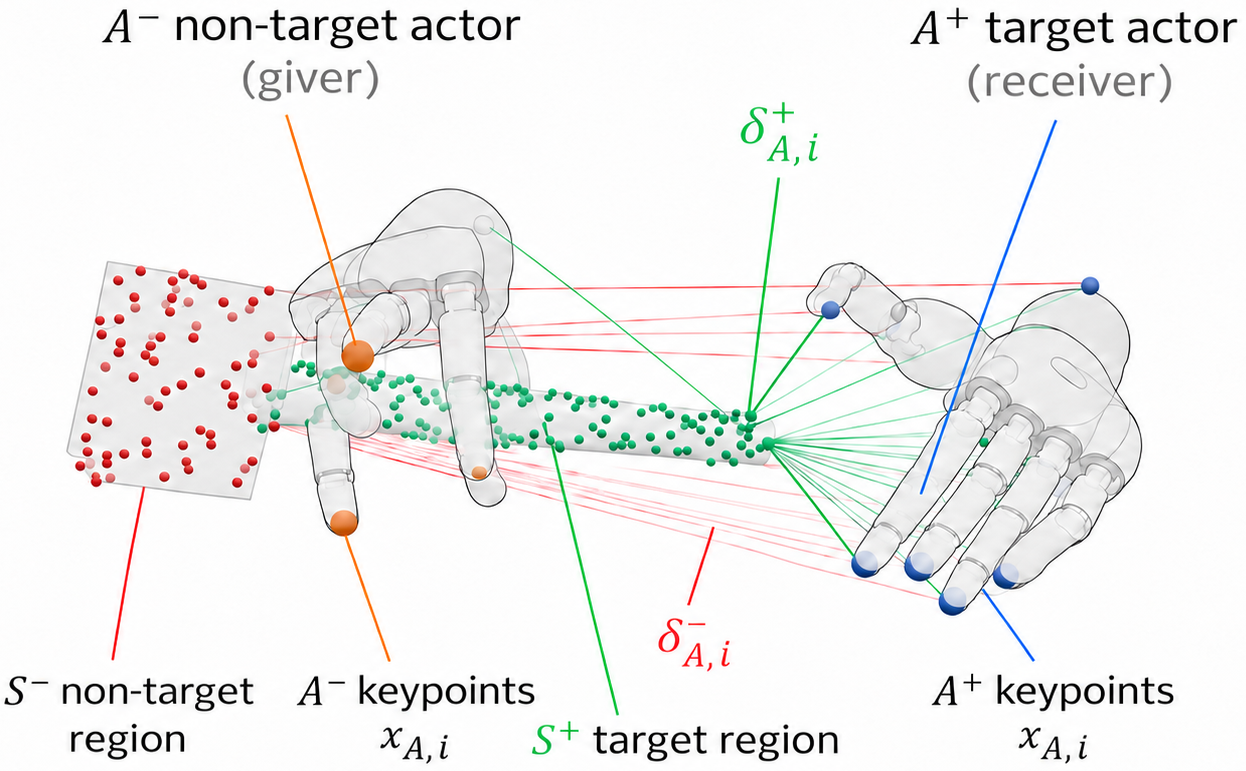}
\caption{\small\textbf{The object-centric representation} This figure shows the representation for the manipulated object duing handover. The object's surface splits into the target contact region $\mathcal{S}^{+}$ (handle, green) and the non-target contact region $\mathcal{S}^{-}$ (head, red); nearest-region displacements $\boldsymbol{\delta}_{A,i}^{+}$ (green) and $\boldsymbol{\delta}_{A,i}^{-}$ (red) relate the keypoints $\mathbf{x}_{A,i}$ of the target actor $A^{+}$ (receiver, blue) and the non-target actor $A^{-}$ (giver, orange) to both regions. }
\vspace{-0.5em}
\label{fig:representation}
\end{figure}

\paragraph{Role-Conditioned Object-Centric Reward}
We reward goal-pose reaching subject to assigned contact roles. For each object $j$, the target actor $A_j^+$ should establish or maintain contact with $S_j^+$, while the non-target actor $A_j^-$ should release the object. These roles correspond to the active arm and table within a workspace, or the receiving and giving arms during transfer. They can change during manipulation, for example when a handover reverses direction. 

All experts share the reward:
{\small
\begin{equation}
r_t=\left[
\sum_{j\in\mathcal{J}_t}
\left(r_t^{\mathrm{pose},j}+r_t^{\mathrm{con+},j}+r_t^{\mathrm{con-},j}+r_t^{\mathrm{succ},j}\right)-r_t^{\mathrm{vel}}\right]r_t^{\mathrm{pen}},
\label{eq:reward}
\end{equation}
}
where $\mathcal{J}_t$ denotes the active objects. The pose term $r_t^{\mathrm{pose},j}$ and target-contact term $r_t^{\mathrm{con+},j}$ reward progress through the shared function~\cite{kedia2026simtoolreal}:
\begin{equation}
\mathcal{R}(d,d^*)=\max(d^*-d,\,0).
\label{eq:pose-progress}
\end{equation}
For each term, $d$ is the current distance and $d^*$ is the previous minimum. Pose history resets for each new goal; target-contact history resets when actor roles change. These two terms reward only new distance improvements, preventing repeated rewards for static proximity or oscillation.

Following~\cite{kedia2026simtoolreal, petrenko2023dexpbt}, the pose term $r_t^{\mathrm{pose},j}$ uses the maximum distance between $N$ corresponding object-frame keypoints, whose world positions under the current and target poses are $\pv_{t,\ell}^j$ and $\hat{\bm{p}}_{t,\ell}^j$:
\begin{equation}
d_t^{\mathrm{pose},j}
=\max_{\ell=1,\ldots,N}\left\|\pv_{t,\ell}^j-\hat{\bm{p}}_{t,\ell}^j\right\|_2.
\label{eq:pose-distance}
\end{equation}
This metric combines translation and rotation errors without separately tuned weights.

$r_t^{\mathrm{con+},j}$ encourages $A_j^+$ to approach the target region $S_j^+$, while $r_t^{\mathrm{con-},j}$ encourages separation of $A_j^-$ from the full object surface $S_j=S_j^+\cup S_j^-$. For hand actors, the corresponding distances are
\begin{equation}
d_t^{\mathrm{con+},j}=\sum_i\left\|\deltav_{A_j^+,i}^{+}\right\|_2,
\qquad
d_t^{\mathrm{con-},j}=\min_i\left\|\deltav_{A_j^-,i}\right\|_2.
\label{eq:target-contact-distance}
\end{equation}
The separation reward $r_t^{\mathrm{con-},j}$ includes a bonus $B_{\mathrm{transfer}}$ awarded once when actor roles change. $r_t^{\mathrm{succ},j}$ rewards reaching each goal pose while satisfying its contact requirements. Finally, $r_t^{\mathrm{vel}}$ penalizes joint velocities, and the safety factor $r_t^{\mathrm{pen}}$ scales down the reward when the two arms and hands get too close. Further details appear in Appendix~\ref{app:object-centric-formulation}.

\subsubsection{RL Training Details}
\label{sec:method-rl-training}

We train each expert using EPO~\cite{wang2025evolutionary} with an asymmetric critic. The actor receives only observations available during real-world deployment, while the critic additionally receives privileged simulator state. Following SimToolReal~\cite{kedia2026simtoolreal}, we train on two-part primitive objects with randomized geometry and inertial properties. The body and head surfaces define the target region $S_j^+$ and avoidance region $S_j^-$, respectively. We also adopt its success tolerance curriculum and sim-to-real domain randomization. Specifically, we randomize robot and object dynamics, inject observation noise, and apply external force and torque perturbations to the objects.

\subsection{Distilling Workspace Experts into a Unified Policy}
\label{sec:method-distillation}

We combine the three experts' complementary behaviors in a unified bimanual policy $\pi_U$, using the same object-centric observations and action space as $\pi_H$. To meet the requirements of diverse downstream tasks, we design three training scene distributions covering expert behaviors and their composition, with routing rules that select the appropriate experts. Since switching experts can expose them to unfamiliar states, we first refine them on transition states collected in end-to-end cross-workspace scenes. The refined experts then supervise $\pi_U$ through DAgger, alongside reinforcement learning.

\subsubsection{Training Scenes and Expert Routing}

\paragraph{Training Scenes}
All scenes contain two objects, initially one in each workspace.
For objects not assigned for manipulation, we set their goals to their current poses.
At each episode reset, we sample one of the following three scene types with probabilities $0.4$, $0.4$, and $0.2$, respectively:

\textbf{Cross-workspace expert scenes} begin with one grasped object whose goals shift between workspaces, following $\pi_H$'s training setup (Sec.~\ref{sec:cross-workspace-goal}). The other object rests on the table, unassigned for manipulation.

\textbf{End-to-end cross-workspace scenes} begin with both objects resting on the table. One is randomly selected for source-side pickup and manipulation, cross-workspace transfer, and destination-side manipulation. Goals follow the cross-workspace schedule (Sec.~\ref{sec:method-goal-sampling}), with $N_{\mathrm{shift}}$ sampled uniformly from $\{3,\ldots,7\}$ at each reset and $N_{\mathrm{dst}}=2$. The other object remains unassigned for manipulation.

\textbf{Concurrent manipulation scenes} begin with both objects resting on the table. Both arms manipulate simultaneously, each acting on the object in its own workspace. Each object's goal advances independently upon completion.

\paragraph{Expert Routing}

For each object assigned for manipulation, we query the corresponding workspace expert, $\pi_L$ or $\pi_R$, when its current and goal positions lie in the same workspace, and $\pi_H$ when a grasped object must reach a goal in the other workspace. This rule switches teachers as end-to-end scenes alternate between within-workspace manipulation and cross-workspace transfer. For concurrent manipulation, we combine the arm-specific actions from $\pi_L$ and $\pi_R$ into a bimanual action. Expert routing is used only during training and is not exposed to $\pi_U$.

\subsubsection{Transition-Aware Expert Refinement}
\label{sec:method-skill-refinement}

During rollouts in end-to-end cross-workspace scenes, we observe excessive and unstable adjustments when experts encounter unfamiliar states generated by their predecessors. We collect the states at expert transitions into a pool. For each target expert, we sample transition states from its valid predecessors to initialize episodes in its own training scenes. This adapts each expert to states encountered at behavior transitions. The refined experts serve as teachers for distillation.

\subsubsection{Distillation with DAgger and RL}
\label{sec:method-distillation-objective}
During student rollouts, DAgger queries the routed teachers on states visited by $\pi_U$. We optimize
\begin{equation}
\mathcal{L}=w_{\mathrm{DAgger}}\mathcal{L}_{\mathrm{DAgger}}+w_{\mathrm{rl}}\mathcal{L}_{\mathrm{rl}},
\label{eq:distill-loss}
\end{equation}

where $\mathcal{L}_{\mathrm{DAgger}}$ matches the unified policy's actions to teacher actions, and $\mathcal{L}_{\mathrm{rl}}$ is the policy-gradient loss under the shared reward from Sec.~\ref{sec:method-atomic-skills}. The imitation term preserves the experts' established behaviors, while reinforcement learning allows the unified policy to recover from its own errors and improve behavior composition. The loss weights follow the scheduling strategy of~\cite{wu2026perceptive}.

\subsection{Task Generalization through Object Pose Commands}
\label{sec:method-prompting}

At deployment, tasks are specified by object-pose goals that describe the desired object motion. Given object assets and contact regions, our unified command interface accepts these goals from diverse sources. We demonstrate two: \textbf{1) VR:} An operator moves virtual objects using a Meta Quest 3, streaming poses at 60\,Hz to update the active goals. \textbf{2) Human Video:} We use FoundationPose~\cite{wen2024foundationpose} to extract object trajectories from human demonstrations and downsample them into waypoints $\tau^j=\{\hat{\bm{T}}_k^j\}_{k=1}^{K_j}$. Each waypoint becomes the active goal until the completion condition is met (Sec.~\ref{sec:method-problem-formulation}), after which the policy advances to the next waypoint.

The interface supports tasks involving one or two objects. For single-object scenes, we complete the observation with a dummy object whose goal is set to its current pose. For two-object waypoint trajectories, goals advance independently for concurrent manipulation or synchronously for paired motions such as alignment and insertion.

The unified policy executes these object-pose commands through pre-grasp manipulation, grasping, in-hand adjustment, and handover, enabling zero-shot execution of diverse tasks without task-specific fine-tuning or robot-level commands.

\section{Experiment}
\label{sec:experiments}

\subsection{Experimental Setup}
\label{sec:exp-setting}
\begin{figure}[t]
\centering
\includegraphics[width=\columnwidth]{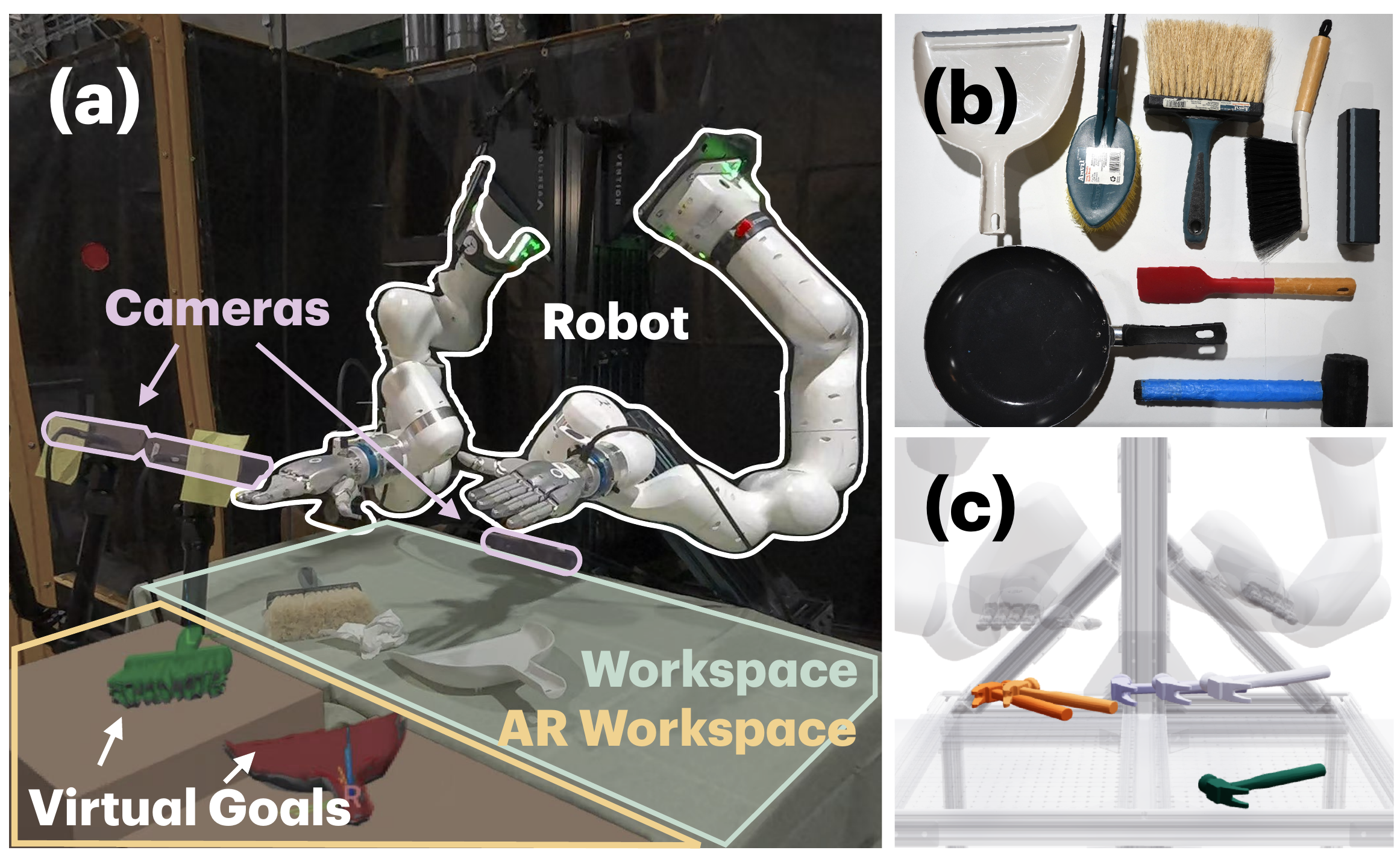}
\caption{\small\textbf{Experimental setup} (a) Real-world setup from the VR view. Three calibrated RGB-D cameras (orange) surround the workspace, and the human controls the object goal poses, shown as rendered object copies, which are streamed to the policy. (b) The real-world object set. (c) An example rule-based object trajectory for the simulated barcode-scanning benchmark.}
\label{fig:setup}
\end{figure}

We evaluate \ours on four real-world tasks and through controlled simulation experiments. The same policy runs at 60\,Hz across all tasks without task-specific fine-tuning. Real-world objects are unseen during training (Fig.~\ref{fig:setup}b) and initialized ungrasped at randomized poses.

\textbf{Hardware Setup}
We use two Franka Research 3 arms with 22-DoF Sharpa Wave hands. Three RGB-D ZED cameras are calibrated to a shared world frame and jointly cover the bimanual workspace (Fig.~\ref{fig:setup}a). A Meta Quest 3 provides object goal poses at 60\,Hz.

\textbf{Object assets and pose tracking} We obtain object meshes using AR Code\footnote{\url{https://ar-code.com/}} on an iPhone or SAM3D~\cite{chen2026sam} from RGB videos. We use SAM 2 with user prompts to segment the body and head regions, defining $S_j^+$ and $S_j^-$, respectively. To track two objects at 60\,Hz under heavy occlusion, we develop a multi-view perception pipeline using FoundationPose~\cite{wen2024foundationpose} and Cutie~\cite{cheng2023putting} on each camera. Per-view pose estimates are transformed into the shared world frame and filtered by observed-to-rendered depth consistency. Failed trackers are reinitialized from reliable estimates in other views, while the remaining poses are fused using Fixed Lag Smoothing (FLS) over an $\mathrm{SE}(3)$ pose graph~\cite{chen2026ptld}. The policy uses the latest pose estimates at each control step.

\subsection{Real-World Evaluation}
\label{sec:exp-results}

\begin{figure*}[t!]
\centering
\includegraphics[width=0.97\textwidth]{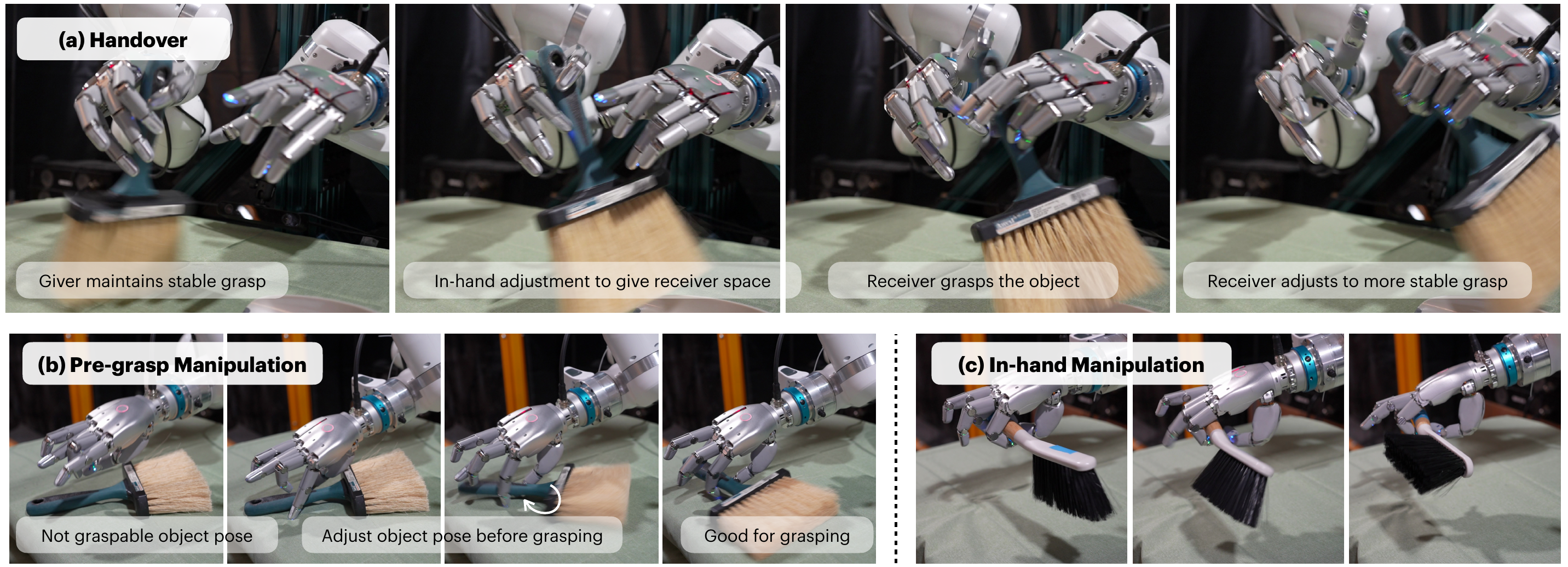}
\caption{\small\textbf{Emergent behaviors on real hardware} (a) Handover: the giver holds a stable grasp, adjusts the brush in-hand to expose a graspable region for the receiver, and releases once the receiver has grasped and settled into a more stable grasp. (b) Pre-grasp manipulation: when the object rests in a pose that is not graspable, the hand first pushes it into a graspable pose. (c) In-hand manipulation: the hand reorients the brush within the grasp to track subsequent object goals. None of these behaviors is specified explicitly; the policy discovers them from object-pose goals alone.}
\label{fig:qualitative}
\end{figure*}

\subsubsection{Tasks and Evaluation Protocol}
We evaluate our unified \ours policy on four real-world tasks without task-specific fine-tuning. Together, these tasks cover complementary dimensions of bimanual dexterity, including object handover, coordinated manipulation of two objects, and precise object alignment. We conduct 20 trials per task using unseen objects initialized at randomized poses. When both objects initialized in same workspace, handover is required. 

\textbf{Barcode Scanning}
We conduct trials across three distinct brushes, each initialized in either the scanner-side or opposite workspace. The robot must grasp and orient the brush to align its barcode with a fixed scanner, transferring it between hands when necessary. The task require precise alignment between barcode and scanner. A trial succeeds when the scanner detects the barcode.

\textbf{Peg insertion}
A peg and a holed block are initialized independently in either workspace or the same workspace. The robot must coordinate both hands to align and insert the peg. Success requires reaching the specified insertion depth.

\textbf{Cleaning}
A brush and a dustpan are initialized independently in the same or separate workspaces. The robot must acquire both tools and execute a coordinated sweep while maintaining tool alignment. Success requires completing the sweep twice.

\textbf{Cooking}
A spatula and a frying pan are initialized in separate workspaces. The robot must stabilize the pan with one hand while executing a stirring motion with the spatula in the other. Success requires completing two cycles of stirring motions.

\subsubsection{Results}
\begin{table}[t]
\centering
\caption{\small \textbf{Real-world results on four tasks combining complementary bimanual capabilities.} The same \ours policy is
used for all tasks without task-specific fine-tuning. Success is
reported as the number of successful trials out of 20.
}
\label{tab:real_world}

\footnotesize
\setlength{\tabcolsep}{1.5pt}
\renewcommand{\arraystretch}{1.2}

\begin{tabularx}{\columnwidth}{
  @{}
  >{\raggedright\arraybackslash}X
  >{\centering\arraybackslash}m{1.3cm}
  >{\centering\arraybackslash}m{1.6cm}
  >{\centering\arraybackslash}m{1.0cm}
  >{\centering\arraybackslash}m{1.05cm}
  @{}
}

\toprule

Task
& Handover
& \makecell{Two-Object\\Coordination}
& Precision
& SR\\
\midrule

Barcode scanning
& \greencheck &            & \greencheck  & 15/20 \\

Peg insertion
& \greencheck & \greencheck & \greencheck  & 15/20 \\

Cleaning
& \greencheck & \greencheck &             & 16/20 \\

Cooking
&            & \greencheck &             & 15/20 \\

\midrule
Average (\%)
& & & & 76.25 \\

\bottomrule
\end{tabularx}
\end{table}

Table~\ref{tab:real_world} shows that \ours achieves an overall success rate of 76.3\% with similar performance across four real-world tasks. Trained only on primitive objects, the same policy generalizes to both novel tasks and unseen real-world objects without task-specific fine-tuning.

Fig.~\ref{fig:qualitative} illustrates the emergent dexterity exhibited by \ours. During handover, the giver reorients the object to expose an accessible region, the receiver establishes a stable grasp, and the giver releases. The policy also performs pre-grasp manipulation to reposition objects from ungraspable configurations, and in-hand manipulation to reach poses unavailable through arm motion alone because of constraints from the table or the other arm. These arm and hand behaviors are not specified by the operator, but arise autonomously as \ours tracks the commanded object motion.

\subsection{Simulation Analysis and Ablation Study}
\begin{table}[t]
  \centering
  \caption{\small \textbf{Simulation results on barcode scanning.} We compare full \ours policy with baselines and distillation variants. Bold marks the best result per metric.}
  \label{tab:sim_ablation}

  \footnotesize
  \setlength{\tabcolsep}{2.5pt}
  \renewcommand{\arraystretch}{1.12}

  \begin{threeparttable}
  \begin{tabularx}{\columnwidth}{
      @{}
      >{\raggedright\arraybackslash}X
      ccc
      @{}
  }
  \toprule
  Method
  & \makecell{SR (\%)\\$\uparrow$}
  & \makecell{Prog. (\%)\\$\uparrow$}
  & \makecell{Time (s)\\$\downarrow$} \\
  \midrule

  \multicolumn{4}{@{}l}{\textbf{\textit{Chained expert teachers}}} \\[1pt]
  \ours teacher w/o cross-workspace expert
  & 60.4 & 79.6 & 9.2 $\pm$ 4.9 \\

  \ours teacher w/o transition-aware refinement
  & 79.4 & 88.1 & 9.1 $\pm$ 4.4 \\
  \ours teacher
  & 84.3 & 89.8 & 8.7 $\pm$ 4.0 \\

  \midrule
  \multicolumn{4}{@{}l}{\textbf{\textit{Unified student policy}}} \\[1pt]
  \ours w/o expert scenes
  & 68.3 & 84.8 & 8.8 $\pm$ 4.4 \\
  \ours w/o RL loss
  & 81.6 & 89.5 & 9.0 $\pm$ 4.2 \\
  \ours
  & \textbf{85.0}
  & \textbf{90.8}
  & \textbf{8.6 $\pm$ 3.9} \\

  \bottomrule
  \end{tabularx}
  \end{threeparttable}
\end{table}
\begin{figure}[b]
\centering
\vspace{-0.5em}
\includegraphics[width=0.85\columnwidth]{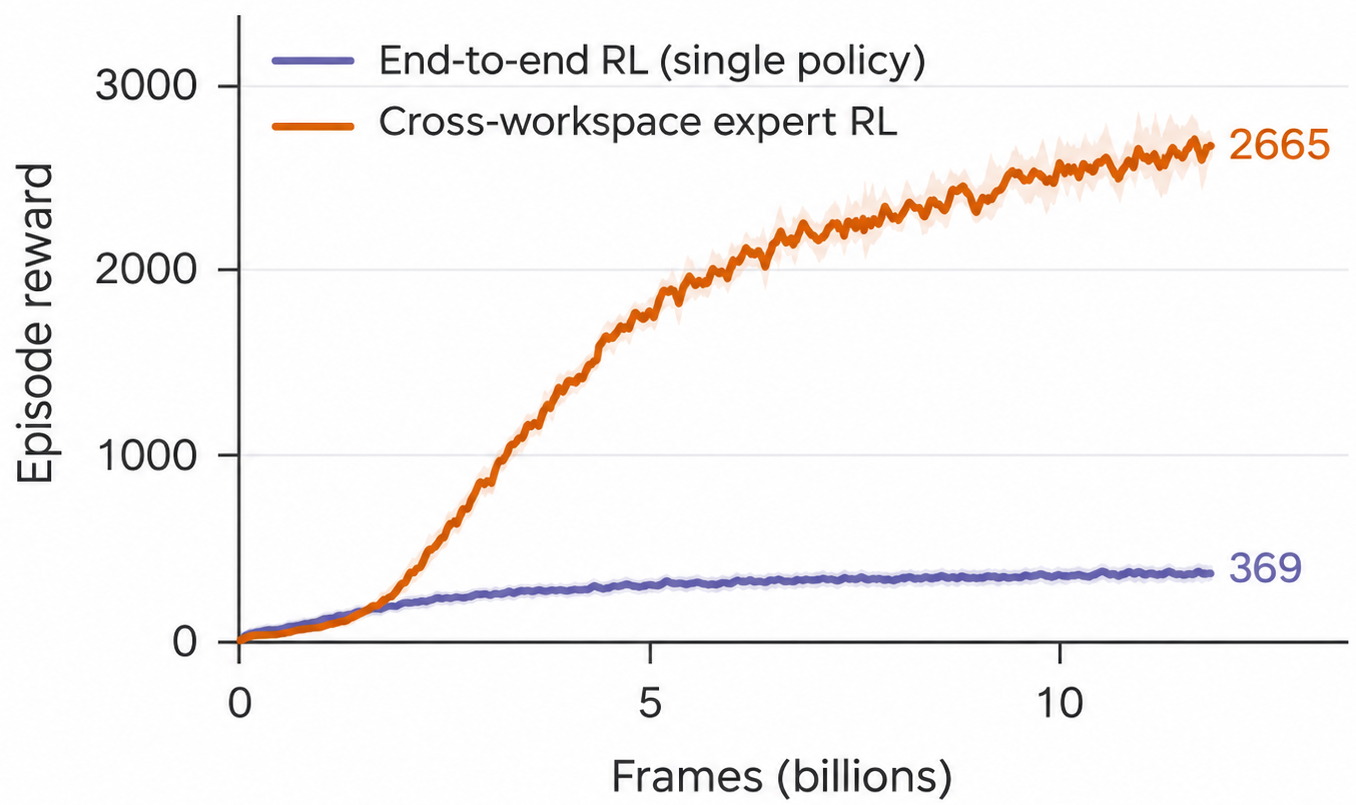}
\caption{\small Expert training vs.\ End-to-End training. 
}
\label{fig:algo_ablation}
\vspace{-0.5em}
\end{figure}

We use barcode scanning as a controlled simulation benchmark (Fig.~\ref{fig:setup}c) that follows rule-based trajectories that interpolates position and orientation linearly, because successful execution requires grasping an initially ungrasped object, reorienting it toward the scanner, transferring it to scanner workspace when necessary, and precisely reaching the final scan pose. We evaluate each method over 3,072 envs each with distinct unseen primitive objects and randomized initial poses across bimanual workspace. Each env contains $K=7$ task waypoints. We report three metrics: \emph{success rate} (SR), the percentage of episodes in which the policy reaches the final task goal; \emph{task progress} (Prog.), the percentage of waypoints reached in each episode, averaged over all episodes; and \emph{completion time} (Time), the elapsed simulator time from episode initialization to the final goal, reported as mean and standard deviation over successful episodes.

\subsubsection{Why train workspace expert separately} 

\ours separately trains three expert policies before unifying them. Fig.~\ref{fig:algo_ablation} compares cross-workspace expert training with an end-to-end baseline trained from scratch on all three workspace regimes, using three seeds and the same algorithm, environment count, initial-state distribution, and compute budget. End-to-end training becomes particularly challenging when all required behaviors must be learned jointly from scratch. This approach also limits extensibility, as incorporating new skills requires retraining on the expanded task distribution. In contrast, the expert-based approach allows new skills to be trained independently and integrated into the unified policy through distillation.

\subsubsection{Why do we need handover} 

We compare with \ours teacher w/o cross-workspace expert, which transfers objects by placing them in the shared workspace for the other hand to regrasp. In contrast, \ours directly coordinates the two hands to transfer the object without an intermediate placement. Adding the handover expert, by chaining all three using a workspace-based routing rule, even without transition-aware refinement (\ours w/o transition-aware refinement), raises success to 79.4\% and progress from 79.6\% to 88.1\%, while reducing completion time. Without handover, failures are dominated by timeouts and drops during placement and regrasping, whereas direct transfer introduces only a small fraction of handover-specific failures.

\subsubsection{Why is transition-aware expert refinement needed} Although rule-based chaining selects the appropriate expert, it does not ensure that the succeeding expert is prepared for the state produced by its predecessor. In particular, the hand configuration produced by one expert may be poorly matched to the preferred manipulation configuration of the next. Despite maintaining a high success rate through the experts' recovery ability and hand-configuration generalization, the policy often performs excessive in-hand adjustment and regrasping. We therefore refine each expert on states
collected at transitions from its predecessors, enabling more efficient and stable transitions between experts. This improves success from 79.4\% to 84.3\% and reduces completion time from 9.1\,s to 8.7\,s.

\subsubsection{How effective is the distillation?}
Removing cross-workspace expert scenes (\ours w/o expert scenes) reduces success to 68.3\%, primarily due to degraded handover performance, indicating that composition scenes alone cannot preserve learned transfer behavior. Removing the RL loss (\ours w/o RL loss) yields 81.6\% success, below the refined teachers' 84.3\%. Adding RL helps the student better capture expert behaviors and surpass the teachers by encouraging preceding behaviors to produce states better suited to subsequent manipulation. The unified policy achieves 85.0\% success.

\section{Conclusion}

We presented \ours, a bimanual policy that learns dexterous arm--hand coordination from a shared object-centric objective rather than behavior-specific rewards. By varying only the goal-sampling regime under one object-centric representation and reward, we train two within-workspace experts and one cross-workspace expert in which grasping, in-hand adjustment, regrasping, and handover---including the \emph{transfer grasps} that expose the region the receiver needs---\emph{emerge} without prescribed motion sequences. We then distill them into a unified policy that is prompted purely with object motion from VR or human video. The single policy succeeds in 61 of 80 real-world trials (76.3\%) on unseen objects across handover, two-object coordination, and precise alignment, and reaches 85.0\% in simulation while matching its refined teachers, suggesting that object-centric task specification offers a simple and scalable interface that shifts the burden from designing how the hands should coordinate to specifying what should happen to the object. 

Limitations remain: \ours requires accurate object meshes and multi-view pose tracking; its 6-DoF goal interface does not directly capture force- or timing-dependent behaviors, motivating richer contact specifications; and goal trajectories are externally provided rather than generated autonomously

\section*{Acknowledgment and AI-Use Disclosure} 
OpenAI Codex was used for minor manuscript editing and code refinement. All scientific content and results were developed and verified by the authors.

\bibliographystyle{IEEEtran}
\bibliography{references}

\section{Appendix}

\subsection{Object-Centric Formulation}
\label{app:object-centric-formulation}
\subsubsection{Role-Conditioned Object-Centric Reward}
\paragraph{Pose progress.}
Following SimToolReal and DexPBT~\cite{kedia2026simtoolreal,petrenko2023dexpbt}, we measure pose error using fixed object-frame keypoints, with placement and scale as in SimToolReal. The pose reward $r_t^{\mathrm{pose},j}$ uses Eq.~\ref{eq:pose-distance}:
\begin{equation}
r_t^{\mathrm{pose},j}=\lambda_{\mathrm{pose}}\max\left(d_t^{\mathrm{pose},j,*}-d_t^{\mathrm{pose},j},0\right),
\label{eq:pose-reward-full}
\end{equation}
Here, $d_t^{\mathrm{pose},j,*}$ is the smallest distance achieved for the current goal and resets when a new goal is sampled.

\paragraph{Target contact.}
The target-contact reward $r_t^{\mathrm{con+},j}$ uses the target-region distance in Eq.~\ref{eq:target-contact-distance}:
\begin{equation}
r_t^{\mathrm{con+},j}=\lambda_{\mathrm{con+}}\max\left(d_t^{\mathrm{con+},j,*}-d_t^{\mathrm{con+},j},0\right),
\label{eq:target-contact-reward-full}
\end{equation}
Here, $d_t^{\mathrm{con+},j,*}$ is the smallest distance achieved under the current actor roles and resets when those roles change.

\paragraph{Non-target separation.}
The separation reward $r_t^{\mathrm{con-},j}$ rewards increases over the initial separation distance:
\begin{equation}
r_t^{\mathrm{con-},j}
=\lambda_{\mathrm{sep}}\max\left(d_t^{\mathrm{con-},j}-d_{\mathrm{init}}^{\mathrm{con-},j},0\right)
+B_{\mathrm{transfer}}\mathbb{I}[\mathcal{T}_t^j].
\label{eq:separation-reward-full}
\end{equation}
Here, $d_{\mathrm{init}}^{\mathrm{con-},j}$ is the initial separation distance under the current actor roles and resets when those roles change. Each role change $\mathcal{T}_t^j$ from goal sampling earns $B_{\mathrm{transfer}}$ once.
For a giving hand, $d_t^{\mathrm{con-},j}$ is defined in Eq.~\ref{eq:target-contact-distance}. For the table, we use the vertical distance from the object-pose origin to the tabletop:
\begin{equation}
d_t^{\mathrm{con-},j}=z_t^j-z_{\mathrm{table}},
\qquad A_j^-=\mathrm{table},
\label{eq:table-separation}
\end{equation}
where $z_t^j$ is the object's vertical position and $z_{\mathrm{table}}$ is the tabletop height. With a fixed table height, the dense term reduces to $\lambda_{\mathrm{sep}}\max(z_t^j-z_{\mathrm{init}}^j,0)$.

\paragraph{Goal completion.}
The goal-pose success reward $r_t^{\mathrm{succ},j}$ is
\begin{equation}
r_t^{\mathrm{succ},j}=B_{\mathrm{succ}}\mathbb{I}\!\left[d_t^{\mathrm{pose},j}<\epsilon\right]\mathbb{I}[\mathcal{C}_t^j],
\label{eq:success-reward}
\end{equation}
where $\epsilon$ is the pose tolerance. Upon success, $B_{\mathrm{succ}}$ is awarded once and a new goal is sampled. The condition $\mathcal{C}_t^j$ is always true within a workspace; during transfer, it requires contact by $A_j^+$ and release by $A_j^-$.

\paragraph{Motion regularization.}
The joint-velocity penalty $r_t^{\mathrm{vel}}$ is
\begin{equation}
r_t^{\mathrm{vel}}
=\sum_{A\in\mathcal{A}_t}\left(
\lambda_{\mathrm{arm}}\|\dot{\qv}_{A,t}^{\mathrm{arm}}\|_1
+\lambda_{\mathrm{hand}}\|\dot{\qv}_{A,t}^{\mathrm{hand}}\|_1
\right),
\label{eq:velocity-reward}
\end{equation}
where $\mathcal{A}_t$ contains the active arms, $\dot{\qv}_{A,t}^{\mathrm{arm}}$ and $\dot{\qv}_{A,t}^{\mathrm{hand}}$ are arm and hand joint velocities.

\paragraph{Arm and hand proximity.}
Let $\pv_{L,i}$ and $\pv_{R,k}$ denote keypoint positions on the left and right arms and hands. Their minimum distance $d_t^{\mathrm{arm}}$ determines the safety factor $r_t^{\mathrm{pen}}$:
\begin{equation}
\begin{aligned}
d_t^{\mathrm{arm}}
&=\min_{i,k}\|\pv_{L,i}-\pv_{R,k}\|_2,\\
r_t^{\mathrm{pen}}
&=r_{\min}+(1-r_{\min})\left(1-\exp(w_{\mathrm{pen}}d_t^{\mathrm{arm}})\right).
\end{aligned}
\label{eq:safety-penalty}
\end{equation}
With $w_{\mathrm{pen}}<0$ and $0<r_{\min}<1$, $r_t^{\mathrm{pen}}$ increases from $r_{\min}$ toward one as clearance increases, attenuating positive rewards at small clearances.

\end{document}